\documentclass[10pt,twocolumn,letterpaper]{article}

\usepackage{cvpr}              

\usepackage{graphicx}
\usepackage{amsmath}
\usepackage{amssymb}
\usepackage{booktabs}
\usepackage{setspace}
\usepackage{longtable}
\usepackage{multirow}
\usepackage{multirow}
\usepackage{amssymb}
\usepackage{pifont}

\usepackage{setspace}
\usepackage{array}
\usepackage{threeparttable}
\usepackage{bm}
\usepackage{caption}
\usepackage{nicefrac}
\usepackage{mathrsfs}
\usepackage{enumitem}
\usepackage{colortbl}
\usepackage{multirow}
\usepackage{marvosym}
\usepackage[accsupp]{axessibility}

\usepackage[export]{adjustbox}

\usepackage[pagebackref,breaklinks,colorlinks]{hyperref}

\usepackage[capitalize]{cleveref}
\crefname{section}{Sec.}{Secs.}
\Crefname{section}{Section}{Sections}
\Crefname{table}{Table}{Tables}
\crefname{table}{Tab.}{Tabs.}

\usepackage{arydshln}

\usepackage{color}
\usepackage[dvipsnames]{xcolor}
\newcommand{\xmarkg}{\textcolor{lightgray}{\ding{55}}\xspace}%

\newcommand{\red}[1]{\textcolor{red}{#1}}
\newcommand{\blue}[1]{\textcolor{blue}{#1}}

\def\cvprPaperID{285} 
\def\confName{CVPR}
\def\confYear{2023}

\begin{document}

\title{Omni Aggregation Networks for Lightweight Image Super-Resolution}

\author{
	Hang Wang\textsuperscript{\rm 2}\footnotemark[1] \quad
	Xuanhong Chen\textsuperscript{\rm 1}\thanks{Equal Contribution.} \quad
	Bingbing Ni\textsuperscript{\rm 1,2}\thanks{Corresponding author: Bingbing Ni.} \quad
	Yutian Liu\textsuperscript{\rm 1} \quad 
	Jinfan Liu\textsuperscript{\rm 1} \quad
        \\
	\textsuperscript{\rm 1}Shanghai Jiao Tong University, Shanghai 200240, China  \quad
	\textsuperscript{\rm 2} Huawei HiSilicon 
   \\
 	francis970625@gmail.com  \,
	\{chen19910528,nibingbing,stau7001,ljflnjz\}@sjtu.edu.cn
}

\maketitle


\begin{abstract}

While lightweight ViT framework has made tremendous progress in image super-resolution, its uni-dimensional self-attention modeling, as well as homogeneous aggregation scheme, limit its effective receptive field (ERF) to include more comprehensive interactions from both spatial and channel dimensions.
To tackle these drawbacks, this work proposes two enhanced components under a new Omni-SR architecture. First, an Omni Self-Attention (OSA) block is proposed based on dense interaction principle, which can simultaneously model pixel-interaction from both spatial and channel dimensions, mining the potential correlations across omni-axis (i.e., spatial and channel). Coupling with mainstream window partitioning strategies, OSA can achieve superior performance with compelling computational budgets.
Second, a multi-scale interaction scheme is proposed to mitigate sub-optimal ERF (i.e., premature saturation) in shallow models, which facilitates local propagation and meso-/global-scale interactions, rendering an omni-scale aggregation building block.
Extensive experiments demonstrate that Omni-SR achieves record-high performance on lightweight super-resolution benchmarks (e.g., $\textbf{26.95}dB$@Urban100 $\times 4$ with only $\textbf{792}$K parameters). Our code is available at \url{https://github.com/Francis0625/Omni-SR}.

\end{abstract}

\section{Introduction}
\label{sec:introduction}
Image super-resolution (SR) is a long-standing low-level problem that aims to recover high-resolution (HR) images from degraded low-resolution (LR) inputs.
Recently, vision transformer~\cite{vaswani2017attention,vision_transformer} based (i.e., ViT-based) SR frameworks~\cite{ipt,swinir} have emerged, showing significant performance gains compared to previously dominant Convolutional Neural Networks (CNNs)~\cite{zhang2018image}.
However, most attempts~\cite{swinir} are devoted to improving the large-scale ViT-based models, while the development of lightweight ViTs (typically, less than $1$M parameters) remains fraught with difficulties.
This paper focuses on boosting the restoration performance of lightweight ViT-based frameworks.

\begin{figure}[t]	
    \centering	
	\includegraphics[width=1.0\linewidth]{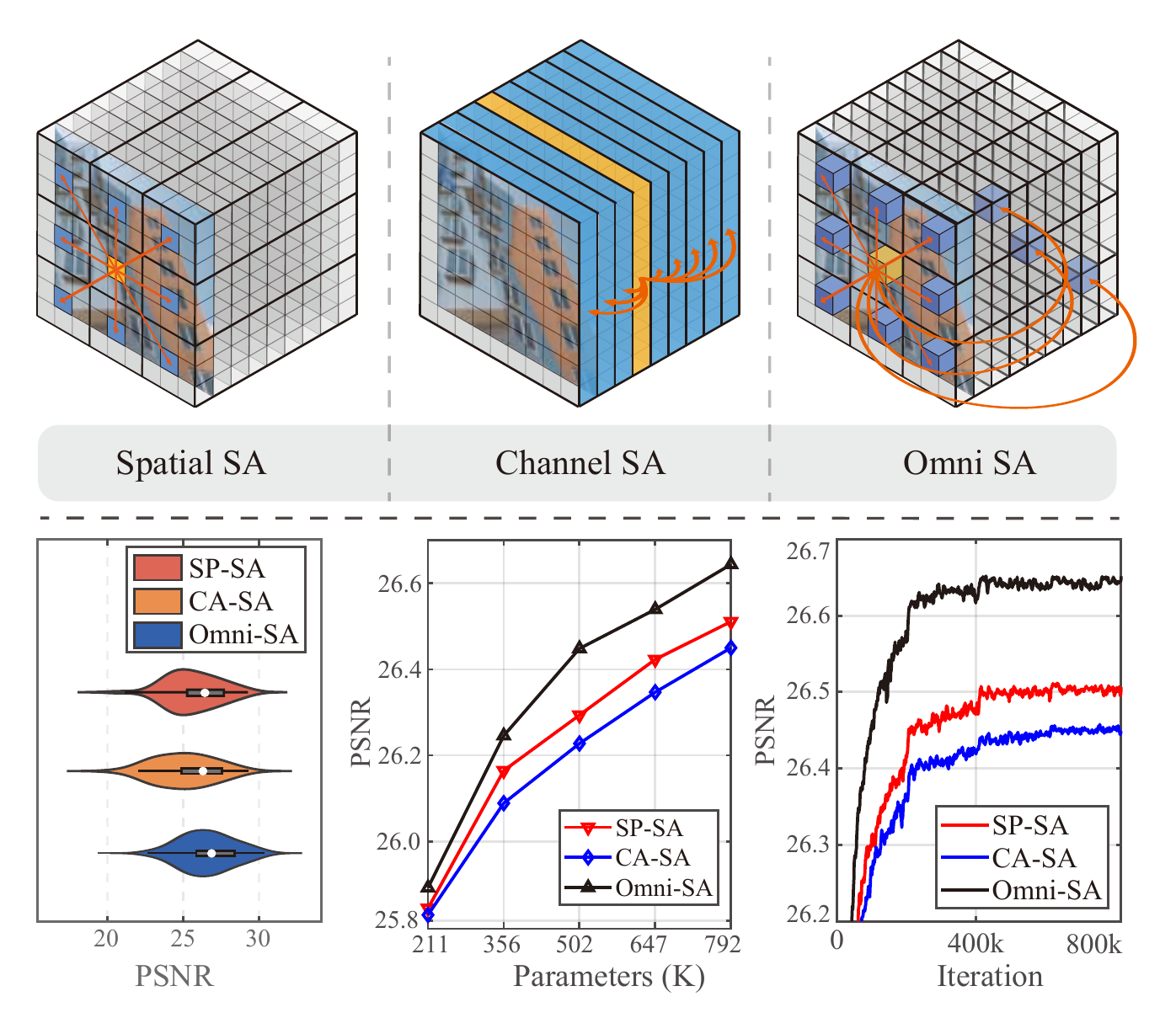}
    \vspace{-6mm}
	\caption{ 
    Typical self-attention schemes~\cite{swinir,restormer} can only perform uni-dimensional (e.g., spatial-only) interactions, and.
	}
    \label{fig:motivation}
    \vspace{-4mm}
\end{figure}


Two difficulties hinder the development of lightweight ViT-based models:
1) \textbf{Uni-dimensional aggregation operators} (i.e., spatial~\cite{swinir} only or channel~\cite{restormer} only) imprisons the full potential of self-attention operators.
Contemporary self-attention generally realizes the interaction between pixels by calculating the cross-covariance of the spatial direction (i.e., width and height) and exchanges context information in a channel-separated manner.
This interaction scheme ignores the explicit use of channel information.
However, recent evidences~\cite{restormer} and our practice show that self-attention in the channel dimension (i.e., computationally more compact than spatial self-attention) is also crucial in low-level tasks.
2) \textbf{Homogeneous aggregation schemes} (i.e., Simple hierarchical stacking of single operators, e.g., convolution, self-attention) neglect abundant texture patterns of multi-scales, which is urgently needed in SR task.
Specifically, a single operator is only sensitive to information of one scale~\cite{largekernel, CP2016Deeplab}, e.g., self-attention is sensitive to long-term information and pays little attention to local information.
Additionally, stacking of homogeneous operators proves to be inefficient and suffers from premature saturation of the interaction range~\cite{HAT}, which is reflected as a suboptimal effective receptive field.
The above problem is exacerbated in lightweight models because lightweight models cannot stack enough layers.

In order to solve the above problems and pursue higher performance, this work proposes a novel omni-dimension feature aggregation scheme called~\emph{Omni Self-Attention} (OSA) exploiting both spatial and channel axis information in a simultaneous manner (i.e., extends the interaction into three-dimensional space), which offers higher-order receptive field information, as shown in Figure~\ref{fig:motivation}.
Unlike scalar-based (a group of important coefficient) channel interaction~\cite{hu2018squeeze}, OSA enables comprehensive information propagation and interaction by cascading computation of the cross-covariance matrices between spatial/channel dimensions.
The proposed OSA module can be plugged into any mainstream self-attention variants (e.g., Swin~\cite{swin}, Halo~\cite{halo}), which provides a finer granularity of important encoding (compared to the vanilla channel attention~\cite{hu2018squeeze}), achieving a perceptible improvement in contextual aggregation capabilities.
Furthermore, a multi-scale hierarchical aggregation block, named \emph{Omni-Scale Aggregation Group} (i.e., OSAG for short), is presented to achieve tailored encoding of varying scales of texture patterns.
Specifically, OSAG builds three cascaded aggregators: local convolution (for local details), meso self-attention (focusing on mid-scale pattern processing), and global self-attention (pursuing global context understanding), rendering an omni-scale (i.e., local-/meso-/global-scale simultaneously) feature extraction capability.
Compared to the homogenized feature extraction schemes~\cite{swinir,RLFN}, our OSAG is able to mine richer information producing features with higher information entropy.
Coupling with the above two designs, we establish a new ViT-based framework for lightweight super-resolution, called \emph{Omni-SR}, which exhibits superior restoration performance as well as covers a larger interaction range while maintaining an attractive model size, i.e., $792K$.

We extensively experiment with the proposed framework with both qualitative and 
quantitative evaluations on mainstream open-source image super-resolution datasets.
It is demonstrated that our framework achieves state-of-the-art performance at the lightweight model scale (e.g., Urban100$\times 4$: $26.95$dB, Manga109$\times 4$: $31.50$dB).
More importantly, compared to existing ViT-based super-solution frameworks, our framework shows superior optimization properties (e.g., convergence speed, smoother loss landscape), which endow our model with better robustness.

\section{Related Works}
\label{sec:related_works}
\textbf{Image Super-resolution.}
CNNs have achieved remarkable success in image SR task. SRCNN~\cite{SRCNN} is the first work to introduce CNNs into the SR field. Many methods~\cite{vdsr,memnet,zhang2018image} employ skip connection to speed up network convergence and improve the reconstruction quality. Channel Attention~\cite{zhang2018image} is also proposed to enhance the representation ability of the SR model.
In order to obtain better reconstruction quality with limited computing resources, several methods~\cite{tai2017image,DBLP:conf/cvpr/HuiWG18,latticenet,muqeet2020multi} explore lightweight architectural design. DRCN~\cite{DBLP:conf/cvpr/KimLL16} utilizes the recursive operation to reduce the number of parameters. DRRN~\cite{tai2017image} introduces global and local residual learning on the basis of DRCN to accelerate training and improve the quality of details. CARN~\cite{CARN} employs cascading mechanism upon a residual network. IMDN~\cite{imdn} proposes an information multi-distillation block to archive better time performance.
Another line of research is to utilize model compression techniques, e.g., knowledge distillation~\cite{DBLP:journals/corr/HintonVD15,DBLP:conf/accv/GaoZL018,DBLP:conf/cvpr/ZhangC0DXW21} and neural architecture search~\cite{DBLP:conf/icpr/Chu0MXL20}) to reduce computing costs.
Recently, a series of transformer-based SR models~\cite{ipt,swinir,HAT,ESRT} emerge with superior performance. Chen\etal~\cite{ipt} develop a pre-trained model for the low-level computer vision task using the transformer architecture. Based on Swin transformer~\cite{swin}, SwinIR~\cite{swinir} proposes a three-stage framework, refreshing the state-of-the-art of SR task.
More recently, some works~\cite{edt,ipt} explore ImageNet pre-training strategy to further enhance SR performance.

\begin{figure}[t]	
    \centering	
	\includegraphics[width=1.0\linewidth]{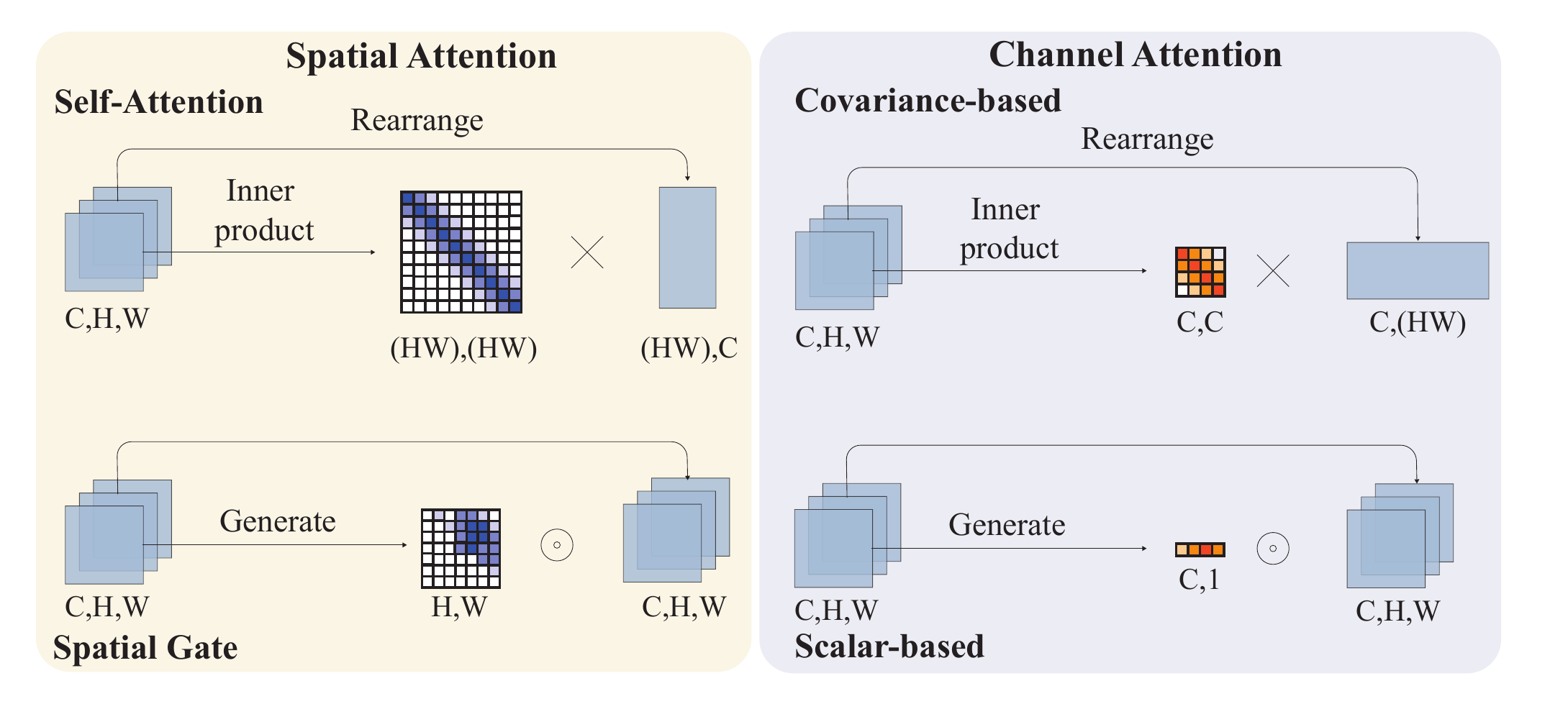}
    \vspace{-5mm}
	\caption{ 
    Illustration of spatial attention and channel attention. These typical attention paradigms only model uni-dimensional (i.e., spatial-only / channel-only) interaction.
	}
    \label{fig:attention}
    \vspace{-4mm}
\end{figure}


\begin{figure*}[t]	
    \centering	
	\includegraphics[width=1.0\linewidth]{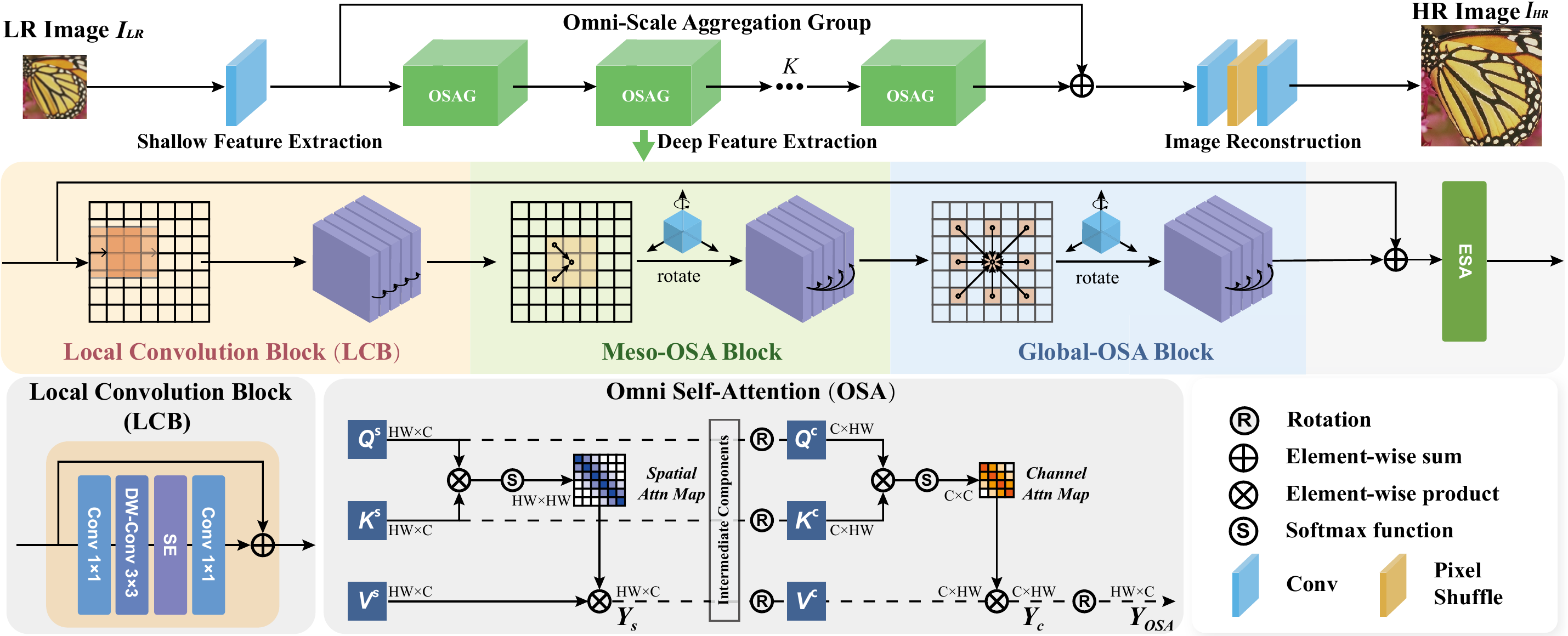}
    \vspace{-6mm}
	\caption{The overall architecture of the proposed Omni-SR framework and structure of OSAG and Omni Self-Attention (OSA).
	}
    \label{fig:framework}
    \vspace{-2mm}
\end{figure*}

\textbf{Lightweight Vision Transformer.}
Due to the urgent demands for applying networks to resource-constrained devices, lightweight vision transformer~\cite{vaswani2017attention,vision_transformer} has attracted widespread attention. Many attempts ~\cite{DBLP:journals/corr/abs-2110-02178,DBLP:conf/cvpr/YangW0ZWLY22,DBLP:journals/corr/abs-2205-03436,nafnet,ESRT,mobileformer,chen2021x,parcnet} have been made to develop lightweight ViTs with comparable performance. A series of methods focus on combining convolutions with transformers to learn both local and global representations. For instance, LVT~\cite{DBLP:conf/cvpr/YangW0ZWLY22} introduces convolution in self-attention to enrich low-level features. MobileViT~\cite{DBLP:journals/corr/abs-2110-02178} replaces matrix multiplication in convolutions with transformer layers to learn global representations. Similarly, EdgeViTs~\cite{DBLP:journals/corr/abs-2205-03436} employs an information exchange bottleneck for full spatial interactions. Different from interpreting convolutions into vision transformers, LightViT~\cite{DBLP:journals/corr/abs-2207-05557} proposes aggregated self-attention for better information aggregation. In this work, we resort to ViT architecture to achieve lightweight and accurate SR.
\section{Methodology} 
\label{sec:methodology}

\subsection{Attention Mechanisms in Super-Resolution}
\label{sec3_1}

Two attention paradigms are widely adopted in SR to assist in analyzing and aggregating comprehensive patterns.

\textbf{Spatial Attention.} Spatial attention can be regarded as an anisotropic selection process.
Spatial self-attention~\cite{vaswani2017attention,ESRT} and spatial gate~\cite{focalnet,nafnet} are predominantly applied.
As shown in Figure~\ref{fig:attention}, spatial self-attention calculates the cross-covariance along the spatial dimension, and the spatial gate generates the channel-separated masks.
Neither of them can transmit information between channels.

\textbf{Channel Attention.}
There are two categories of channel attention, i.e., scalar-based~\cite{hu2018squeeze} and covariance-based~\cite{restormer}, proposed to perform channel
recalibration or transmit patterns among channels.
As shown in Figure~\ref{fig:attention}, the former predicts a group of importance scalars to weigh different channels, while the latter computes a cross-covariance matrix to enable channel re-weighting and information transmission simultaneously.
Compared to spatial attention, channel attention handles spatial dimension isotropically, and thus, the complexity is significantly reduced, which also impairs the accuracy of aggregation.

Several attempts~\cite{woo2018cbam, park2018bam} have proved that both spatial attention and channel attention are beneficial for SR task and their characteristics are complementary to each other, thus integrating them in a computationally compact way would bring notable benefits in expressive capability.

\subsection{Omni Self-Attention Block}
\label{sec3_2}
To mine all the correlations hidden in the latent variables, we propose a novel self-attention paradigm called Omni Self-Attention (OSA) block.
Unlike existing self-attention paradigms (e.g., spatial self-attention~\cite{vaswani2017attention,ipt,ESRT}) that only indulge in unidimensional processing, OSA establishes the spatial and channel context simultaneously. The obtained two-dimensional relationship is highly necessary and beneficial, especially for lightweight models. 
On the one hand, as the network deepens, important information is scattered into different channels~\cite{hu2018squeeze}, and it is critical to deal with them in a timely manner.
On the other hand, although spatial self-attention takes advantage of the channel dimension in calculating the covariance, it does not transmit the information between channels (refer to Sec.~\ref{sec3_1}).
Given the above conditions, our OSA is designed to transmit both spatial and dimensional information in a compact manner.

The proposed OSA calculates the score matrices corresponding to the space and channel direction through sequential matrix operations and rotation, as illustrated in Figure~\ref{fig:framework}. 
Specifically, suppose $X\in \mathbb{R}^{HW\times C}$ denotes the input feature, where $H$ and $W$ are the width and height of the input, and $C$ is the channel number.
Firstly, $X$ is embedded to query,
key and value matrices $Q^s, K^s, V^s \in \mathbb{R}^{HW\times C}$ through linearly projection.
We calculate the production of query and key to get the spatial attention map of size $\mathbb{R}^{HW\times HW}$.
Then we perform the spatial attention to obtain the intermediate aggregated results.
Note that window strategy is usually used to significantly reduce the resource overhead.
Next stage, we \textbf{rotate} the input query and key matrices to get the transposed query and key matrices $Q^c, K^c \in \mathbb{R}^{C \times HW}$, and also \textbf{rotate} the value matrices to get the value matrix $V^c \in \mathbb{R}^{C\times HW}$ for the subsequent channel-wise self-attention. The obtained channel-wise attention map of size $\mathbb{R}^{C\times C}$ models channel-wise relationships. Finally, we get the final aggregated $Y_{\small{OSA}}$ by the inverse rotation of the channel attention output $Y_{c}$.
The whole OSA process is formulated as follows:
{\small
\begin{flalign}
\label{osa}
&Q^s=X\cdot\mathcal{W}_q,\ \ K^s=X\cdot\mathcal{W}_k,\ \
V^s=X\cdot\mathcal{W}_v, \\
&Y_{s} = \mathcal{A}^s(Q^s,K^s,V^s)=\text{SoftMax}(Q^sK^{sT})\cdot V^s, \ \\ \vspace{2mm}
&Q^c=\mathcal{R}(Q'),\quad K^c=\mathcal{R}(K'),\quad
V^c=\mathcal{R}(V'), \\
&Y_{c} = \mathcal{A}^c(Q^c,K^c,V^c)=\text{SoftMax}(K^cQ^{cT})\cdot V^c, \, \\ \vspace{2mm}
&Y_{\small OSA} = \mathcal{R}^{-1}(Y_{c}), 
\end{flalign}
}
where $\mathcal{W}_q, \mathcal{W}_k, \mathcal{W}_v$ denote the linear projection matrices for the query, key, and value, respectively.
$Q',K',V'$ are the input embedding matrices of channel-wise self-attention, which are embedded from fore spatial self-attention or or copied directly from $Q^s,K^s,V^s$.
$\mathcal{R}(\cdot)$ denotes the rotation operation around spatial axis and $\mathcal{R}^{-1} (\cdot)$ is the inverse rotation. Some normalization factors are omitted for the sake of simplicity.
In particular, this design shows compelling properties that can integrate the element-wise results of two matrix operations (i.e., spatial-/channel- matrix operation), thereby enabling omni-axial interactions.
Note that our proposed OSA paradigm can be a drop-in replacement of the Swin~\cite{swin,swinir} attention block to higher performance with less parameters. Benefiting from the smaller attention map size of channel self-attention, the proposed OSA is less computationally intensive compared to the cascade shifted-window self-attention scheme in Swin.


\emph{Discussion with other hybrid attention paradigms.} Compared to previous hybrid channel and spatial attention works like CBAM~\cite{woo2018cbam} and BAM~\cite{park2018bam}, 
their scalar-based attention weights only reflect the relative degree of importance, without further inter-pixel information exchange, leading to limited relation modeling capability. Several recent works~\cite{HAT} also incorporate channel attention with spatial Self-attention, but these attempts only resort to scalar weights for channel recalibration, while our OSA paradigm enables channel-wise interaction to mine the potential correlations in omni-axis. Performance comparison of different attention paradigms can be found in Sec.~\ref{sec_ablation}.

\subsection{Omni-Scale Aggregation Group}
\label{sec3_3}
How to utilize the proposed OSA paradigm to build a high-performance and compact network is another key topic.
Although hierarchical stacking of windows-based self-attention (e.g., swin~\cite{swin,swinir}) has become mainstream, various works have found that the window-based paradigms are very inefficient for large-range interactions, especially for shallow networks.
It is worth pointing out that large-range interaction can provide a pleasing effective receptive field, which is crucial for improving image restoration performance~\cite{ESRT}.
Unfortunately, direct global interaction is resource-prohibitive and detracts from local aggregation capabilities.
Taking these points into account, we propose an Omni-Scale Aggregation Group (i.e., OSAG for short) to
pursue progressive receptive field feature aggregation with low computational complexity.
As shown in Figure~\ref{fig:framework}, OSAG mainly consists of three stages: local, meso and global aggregations.
Specifically, a channel attention~\cite{hu2018squeeze} enhanced inverted bottleneck~\cite{howard2017mobilenets} is introduced to fulfill the local pattern process with limited overhead. Based on the proposed OSA paradigm, we derive two instances (i.e., Meso-OSA and Global-OSA) responsible for the interaction and aggregation of meso and global information. Note that the proposed omni self-attention paradigm can be used for different purposes.
Meso-OSA performs attention within a group of non-overlap patches, which restricts Meso-OSA to only focus on meso-scale pattern understanding.
Global-OSA samples data point sparsely across the entire feature within an atrous manner, endowing Global-OSA with the ability to achieve global interactions at a compelling cost.

\begin{figure}[t]	
    \centering	
	\includegraphics[width=0.9\linewidth]{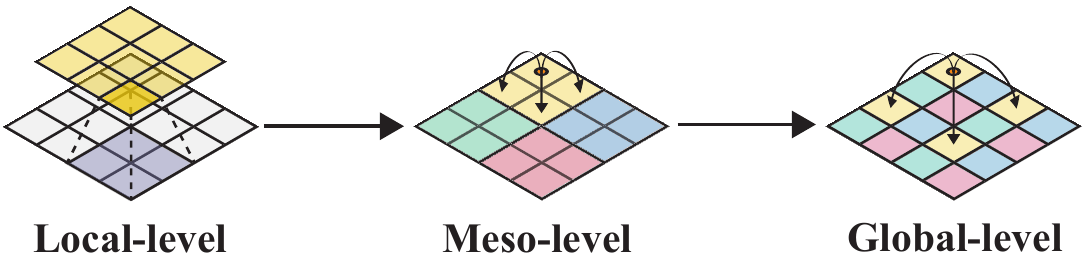}
	\caption{
    Illustration of omni-scale aggregation scheme. Our proposed Omni-SR contains three types of feature aggregation at local level, meso and global level, respectively. 
	}
    \label{fig:interaction}
    \vspace{-4mm}
\end{figure}


The only difference between Meso-OSA and Global-OSA is the window partition strategy, as shown in Figure~\ref{fig:interaction}.
In order to achieve meso-scale interaction, Meso-OSA split the input feature $X$ into non-overlapping blocks with size $P\times P$.
Note that after window partition, the block dimensions are gathered onto the spatial dimension (i.e., -2 axis): {\small $(H,W,C)\rightarrow  (\frac{H}{P}\times P,\frac{W}{P}\times P,C)\rightarrow (\frac{HW}{P^2},P^2,C)$}.
%
%
While the Global-OSA divides the input feature into a uniform $G\times G$ grid, with each lattice having an adaptive size of $\frac{H}{G}\times \frac{W}{G}$. Similar to Meso-OSA, the grid dimension is also gathered on the spatial axis (\ie -2 axis):
{\small $ (H,W,C)\rightarrow(G\times \frac{H}{G},G\times\frac{W}{G},C)$} {\small $ \rightarrow(G^2,\frac{HW}{G^2},C)\rightarrow(\frac{HW}{G^2},G^2,C)$}.

\subsection{Network Architecture}
\label{sec3_4}


\textbf{Overall Structure.}
Based on the Omni Self-Attention paradigm and the Omni-Scale Aggregation Group, we further develop a lightweight Omni-SR framework to achieve high-performance image super-resolution. As shown in Figure~\ref{fig:framework}, Omni-SR consists of three parts, i.e., shallow feature extraction, deep feature extraction, and image reconstruction. Specifically, given the LR input $I_{\textit{LR}} \in \mathbb{R}^{H\times W\times C_{in}}$, we first use 
a $3\times 3$ convolution $H_{\textit{SF}}$ to extract shallow feature $X_0 \in \mathbb{R}^{H\times W\times C}$ as
{\small
\begin{equation}
X_0=H_{\textit{SF}}(I_{\textit{LR}}),
\end{equation}
}
where $C_{in}$ and $C$ denote the channel number of the input and shallow feature. The convolution layer provides a simple way to convert the input from image space into high-dimensional feature space. Then we use $K$ stacked omni-scale aggregation groups (OSAG) and one $3\times 3$ convolution layer $H_{\textit{CONV}}$ in a cascade manner to extract deep feature$F_{\textit{DF}}$. 
Such a process can be expressed as
{\small
\begin{equation}
\begin{split}
X_{i}= H_{\textit{OSAG}_{i}}(X_{i-1}), \quad i=1, 2, \ldots, K, \\
X_{\textit{DF}}=H_{\textit{CONV}}(X_K), \qquad\qquad
\end{split}
\end{equation}
}
where $H_{\textit{OSAG}_{i}}$ represents the $i$-th OSAG, $X_{1}, X_2, \ldots, X_K$ denote intermediate features. Following~\cite{swinir}, we also use a convolutional layer at the end of feature extraction to get better feature aggregation. Finally we reconstruct the HR image $I_{\textit{HR}}$ by aggregating shallow and deep features as
{\small
\begin{equation}
I_{\textit{HR}}=H_{\textit{Rec}}(X_0+X_{\textit{DF}}),
\end{equation}
}
where $H_{\textit{Rec}}(\cdot)$ denotes the reconstruction module. In detail, PixelShuffle~\cite{shi2016real} is used to up-sample the fused feature.
\paragraph{Omni-Scale Aggregation Group (OSAG).} As shown in Figure~\ref{fig:framework}, each OSAG contains a local convolution block (LCB), a meso-OSA block, a global-OSA block, and an ESA block~\cite{RFDB,RLFN}. The whole process can be formulated as
{\small
\begin{gather}
X_{res}=H_{\textit{Global-OSAB}_{i}}(H_{\textit{Meso-OSAB}_{i}}(H_{\textit{LCB}_{i}} (X_{i-1}) )),  \\
X_{i} = H_{\textit{ESA}_{i}} (H_{\textit{Conv}_{i}} (X_{res} + X_{i-1})), 
\end{gather}
}
where $X_{i-1}$ and $X_{i}$ represents the input and output feature of $i$-th OSAG. After the mapping of convolution layers, we insert a Meso-OSAB for window-based self-attention and a Global-OSAB to enlarge the receptive field for better information aggregation. At the end of OSAG, we reserve the convolutional layer and ESA block following~\cite{RLFN,zhang2018image}. 

In specific, LCB is implemented as a stack of pointwise and depthwise convolutions with a CA module~\cite{iandola2016squeezenet} between them to adaptively re-weight channel-wise features. This block aims to aggregate local contextual information as well as to increase the trainability of the network~\cite{XiaoSMDDG21}. Two types of OSA blocks (i.e., Meso-OSA block and Global-OSA block) are then followed to obtain interactions from different regions. Based on different window partition strategies,  Meso-OSA block seeks inner-block interaction, and Global-OSA blocks aim for global mixing.
OSA blocks follow typical Transformers designs with Feedforward network (FFN) and LayerNorm~\cite{layernorm}, 
and the only difference is that the origin self-attention operation is replaced with our proposed OSA operator. For FFN, we adopt the GDFN proposed by Restormer~\cite{restormer}. Combining these individuals seamlessly, the designed OSAG enables information propagation between any pair of tokens in the feature map. We use the ESA module proposed in~\cite{RFDB,RLFN} to further refine the fused feature.




\paragraph{Optimization Objective.}
Following prior works~\cite{edsr,DBLP:conf/cvpr/ZhangTKZ018,swinir,wang2022learning}, we train the model by minimizing a standard $L_1$ loss between model prediction $\hat{I}_{HR}$ and HR label $I_{HR}$ as follows:
\begin{equation} 
\label{loss}
\mathcal{L} = \| I_{\textit{HR}} - {\hat{I}}_{\textit{HR}} \|_1.
\end{equation}

\begin{table*}[t]
  \centering
  \footnotesize
  \setlength{\tabcolsep}{2.3mm}

    \caption{
    Quantitative comparison (PSNR/SSIM) for \textbf{lightweight image SR} with state-of-the-art methods on benchmark datasets. The best and second-best results are marked in \textcolor{red}{red} and \textcolor{blue}{blue} colors, respectively. ``$\dagger$” indicates that model is trained on DF2K.}
  \vspace{-3mm}
  \label{tab:psnr-ssim}
  \renewcommand{\arraystretch}{0.82}
	\begin{tabular}{l|c|c|c|c|c|c|c|c}
		\toprule[1pt]
		\multirow{2}{*}{Method} & \multirow{2}{*}{Years} & \multirow{2}{*}{Scale}  & Params & Set5 & Set14 & BSD100 & Urban100 & Manga109 \\
		&  & & (K) & PSNR / SSIM & PSNR / SSIM & PSNR / SSIM & PSNR / SSIM & PSNR / SSIM \\
    \hline
	VDSR~\cite{vdsr} &   CVPR16 & \multirow{13}{*}{$\times 2$} & 666 & 36.66 / 0.9542 & 33.05 / 0.9127 & 31.90 / 0.8960 & 30.76 / 0.9140 & 37.22 / 0.9750  \\
	
	MemNet~\cite{memnet} &  ICCV17 &  & 678 & 	37.78 / 0.9597 & 33.28 / 0.9142 & 32.08 / 0.8978 & 31.31 / 0.9195 & 37.72 / 0.9740\\
	
	SRMDNF~\cite{srmdnf} &  
	CVPR18 & & 1511 & 37.79 / 0.960 & 33.32 / 0.915 & 32.05 / 0.8985 & 31.33 / 0.9204 & 38.07 / 0.9761\\
	
	CARN~\cite{CARN} &  ECCV18&  & 1,592 & 37.76 / 0.9590 & 33.52 / 0.9166 & 32.09 / 0.8978 & 31.92 / 0.9256  & 38.36 / 0.9765 \\
		
    IMDN~\cite{imdn} &  MM19  && 694 & 38.00 / 0.9605 & 33.63 / 0.9177 & 32.19 / 0.8996 & 32.17 / 0.9283 & 38.88 / 0.9774 \\
    
    RFDN-L~\cite{liu2020residual}  &ECCV20 & & 626 & 38.08 / 0.9606 &  33.67 / 0.9190 & 32.18 / 0.8996 & 32.24 / 0.9290  & 38.95 / 0.9773 \\

    MAFFSRN~\cite{muqeet2020multi}  &ECCV20 && 402 & 37.97 / 0.9603 & 33.49 / 0.9170 & 32.14 / 0.8994 & 31.96 / 0.9268 &  - / -\\
    
    LatticeNet~\cite{latticenet}  &ECCV20  & & 756 & 38.15 / 0.9610 & 33.78 / 0.9193 & 32.25 / 0.9005 & 32.43 / 0.9302 & - / -\\
		
	RLFN~\cite{RLFN} & CVPRW22&   & 527 & 38.07 / 0.9607 & 33.72 / 0.9187 & 32.22 / 0.9000 & 32.33 / 0.9299 & - / -  \\
	
	SwinIR~\cite{swinir} &  ICCVW21& & 878 & 		 38.14 / 0.9611 & 33.86 / 0.9206 & 32.31 / 0.9012 & 32.76 / 0.9340 & 39.12 / 0.9783 \\

	\rowcolor{green!08}
	\textbf{Omni-SR} & Ours &  & 772 &  \textcolor{blue}{38.22} / \textcolor{blue}{0.9613} & \textcolor{blue}{33.98} / \textcolor{blue}{0.9210} & 
 \textcolor{blue}{32.36} / \textcolor{blue}{0.9020} & \textcolor{blue}{33.05} / \textcolor{blue}{0.9363}  & \textcolor{blue}{39.28} / \textcolor{blue}{0.9784} \\
	\rowcolor{green!08}
	\textbf{Omni-SR${\small \dagger}$ } & Ours & & 772  &  \textcolor{red}{\textbf{38.29}} / \textcolor{red}{\textbf{0.9617}} & \textcolor{red}{\textbf{34.27}} / \textcolor{red}{\textbf{0.9238}} & \textcolor{red}{\textbf{32.41}} / \textcolor{red}{\textbf{0.9026}} & \textcolor{red}{\textbf{33.30}} / \textcolor{red}{\textbf{0.9386}}  & \textcolor{red}{\textbf{39.53}} / \textcolor{red}{\textbf{0.9792}} \\
		
	\midrule
	VDSR~\cite{vdsr}  & CVPR16& \multirow{13}{*}{$\times 3$} &   666 & 33.66 / 0.9213 & 29.77 / 0.8314 & 28.82 / 0.7976 & 27.14 / 0.8279 & 32.01 / 0.9340 \\
	
	MemNet~\cite{memnet} & ICCV17 &   & 678 & 34.09 / 0.9248 & 30.00 / 0.8350 & 28.96 / 0.8001 & 27.56 / 0.8376 & 32.51 / 0.9369\\
	
	EDSR~\cite{edsr} &CVPRW17&    & 1,555 & 34.37 / 0.9270 & 30.28 / 0.8417 & 29.09 / 0.8052 & 28.15 / 0.8527 & 33.45 / 0.9439 \\
	
	SRMDNF~\cite{srmdnf} & CVPR18 &   & 1,528 & 34.12 / 0.9254 & 30.04 / 0.8382 & 28.97 / 0.8025 & 27.57 / 0.8398 & 33.00 / 0.9403 \\
	
	CARN~\cite{CARN} & ECCV18&    & 1,592 & 34.29 / 0.9255 & 30.29 / 0.8407 & 29.06 / 0.8034 & 28.06 / 0.8493 & 33.50 / 0.9440 \\
		
    IMDN~\cite{imdn} & MM19 &   & 703 & 34.36 / 0.9270 & 30.32 / 0.8417 & 29.09 / 0.8046 & 28.17 / 0.8519 & 33.61 / 0.9445 \\

    RFDN-L~\cite{liu2020residual} & ECCV20&   & 633 & 34.47 / 0.9280 &  30.35 / 0.8421 & 29.11 / 0.8053 & 28.32 / 0.8547  & 33.78 / 0.9458 \\

    MAFFSRN~\cite{muqeet2020multi} &ECCV20 &  & 807 & 34.45 / 0.9277 & 30.40 / 0.8432 & 29.13 / 0.8061 & 28.26 / 0.8552 &  - / -\\
    
    LatticeNet~\cite{latticenet} &ECCV20 &   & 765 & 34.53 / 0.9281 & 30.39 / 0.8424 & 29.15 / 0.8059 & 28.33 / 0.8538 & - / -\\
		
	ESRT~\cite{ESRT} & CVPRW22 &   & 770 & 34.42 / 0.9268 & 30.43 / 0.8433 & 29.15 / 0.8063 & 28.46 / 0.8574 & 33.95 / 0.9455 \\
	
	SwinIR~\cite{swinir} & ICCVW21 &  & 886  & 		34.62 / 0.9289 & \blue{30.54} / \blue{0.8463}& 29.20 / 0.8082& 28.66 / 0.8624& 33.98 / 0.9478\\
		
	\rowcolor{green!08}
	\textbf{Omni-SR }  & Ours &   & 780 &  \blue{34.70} / \blue{0.9294} & \blue{30.57} / \blue{0.8469} & \blue{29.28} / \blue{0.8094} & \blue{28.84} / \blue{0.8656} & \blue{34.22} / \blue{0.9487} \\

	\rowcolor{green!08}
	\textbf{Omni-SR${\small \dagger}$ } & Ours & & 780 &  \textcolor{red}{\textbf{34.77}} / \textcolor{red}{\textbf{0.9304}} & \textcolor{red}{\textbf{30.70}} / \textcolor{red}{\textbf{0.8489}}  &  \textcolor{red}{\textbf{29.33}} / \textcolor{red}{\textbf{0.8111}} & \textcolor{red}{\textbf{29.12}} / \textcolor{red}{\textbf{0.8712}}  & \textcolor{red}{\textbf{34.64}} / \textcolor{red}{\textbf{0.9507}} \\
		
	\midrule
		
	VDSR~\cite{vdsr} & CVPR16  & \multirow{13}{*}{$\times 4$} & 666 & 31.35 / 0.8838 & 28.01 / 0.7674 & 27.29 / 0.7251 & 25.18 / 0.7524 & 28.83 / 0.8870 \\
	
	MemNet~\cite{memnet} & ICCV17 &    & 678 & 31.74 / 0.8893 & 28.26 / 0.7723 & 27.40 / 0.7281 & 25.50 / 0.7630 & 29.42 / 0.8942 \\
	
	EDSR~\cite{edsr} & CVPRW17 &    & 1,518 & 32.09 / 0.8938 &  28.58 / 0.7813 & 27.57 / 0.7357 & 26.04 / 0.7849 & 30.35 / 0.9067 \\
	
	SRMDNF~\cite{srmdnf} &CVPR18 &   & 1,552 & 31.96 / 0.8925 & 28.35 / 0.7787 & 27.49 / 0.7337 & 25.68 / 0.7731 & 30.09 / 0.9024\\
		
    CARN~\cite{CARN} & ECCV18 &   & 1,592 & 32.13 / 0.8937 & 28.60 / 0.7806 &  27.58 / 0.7349 & 26.07 / 0.7837  & 30.47 / 0.9084 \\
		
    IMDN~\cite{imdn} & MM19 &   & 715 & 32.21 / 0.8948 & 28.58 / 0.7811 & 27.56 / 0.7353 & 26.04 / 0.7838 & 30.45 / 0.9075 \\

    RFDN-L~\cite{liu2020residual} & ECCV20 &   & 643 & 32.28 / 0.8957 & 28.61 / 0.7818 & 27.58 / 0.7363 & 26.20 / 0.7883 & 30.61 / 0.9096 \\

    MAFFSRN~\cite{muqeet2020multi} &ECCV20 &  & 830 & 32.20 / 0.8953 & 26.62 / 0.7822 & 27.59 / 0.7370 & 26.16 / 0.7887 & - / - \\
    
    LatticeNet~\cite{latticenet} & ECCV20 &   & 777 & 32.30 / 0.8962 & 28.68 / 0.7830 & 27.62 / 0.7367 & 26.25 / 0.7873 & - / -\\
	
	RLFN~\cite{RLFN} &  CVPRW22&   & 543 & 32.24 / 0.8952 & 28.62 / 0.7813 & 27.60 / 0.7364 & 26.17 / 0.7877 & - / -  \\
		
	ESRT~\cite{ESRT} & CVPRW22 &   & 751 & 32.19 / 0.8947 & 28.69 / 0.7833 & 27.69 / 0.7379 & 26.39 / 0.7962 & 30.75 / 0.9100 \\
	
	SwinIR~\cite{swinir} & ICCVW21 &   &  897 & 32.44 / 0.8976 & 28.77 / 0.7858& 27.69 / 0.7406 & 26.47 / 0.7980 & 30.92 / 0.9151\\
	
	   \rowcolor{green!08}
	\textbf{Omni-SR} & Ours &  & 792 & \blue{32.49} / \blue{0.8988} & \blue{28.78} / \blue{0.7859}  & \blue{27.71} / \blue{0.7415}  & \blue{26.64} / \blue{0.8018} &  \blue{31.02} / \blue{0.9151} \\
\rowcolor{green!08}
	\textbf{Omni-SR${\small \dagger}$ } & Ours &   & 792 &  \textcolor{red}{\textbf{32.57}} / \textcolor{red}{\textbf{0.8993}} &  \textcolor{red}{\textbf{28.95}} / \textcolor{red}{\textbf{0.7898}}  & \textcolor{red}{\textbf{27.81}} / \textcolor{red}{\textbf{0.7439}}  &  \textcolor{red}{\textbf{26.95}} / \textcolor{red}{\textbf{0.8105}} & \textcolor{red}{\textbf{31.50}} / \textcolor{red}{\textbf{0.9192}} \\
	
	\bottomrule[1pt]
		
  \end{tabular}

\vspace{-0.3cm}
\end{table*}

\section{Experiments}
\label{sec:experiments}
\subsection{Experimental Setup}
\vspace{-4mm}
\quad\par\textbf{Datasets and Metrics.} 
Following previous work~\cite{edsr,timofte2017ntire,latticenet,zhang2018image,swinir}, DIV2K~\cite{timofte2017ntire} and Flickr2K~\cite{timofte2017ntire} are used as training datasets. For a fair comparison, we employ two training protocols, i.e., training with DIV2K only and training with DF2K (DIV2K + Flickr2K). Note that the model trained with DF2K is marked with {small $\dagger$}.
For testing, we adopt five standard benchmark datasets: Set5~\cite{bevilacqua2012low}, Set14~\cite{zeyde2010single},
B100~\cite{martin2001database}, Urban100~\cite{huang2015single} and Manga109~\cite{matsui2017sketch}. 
PSNR and SSIM~\cite{wang2004image} are adopted to evaluate the SR performance on the Y channel of the transformed YCbCr space.

\textbf{Implementation Details.}
During training, we augment the data with random horizontal flips and 90/270-degree rotations. LR images are generated by bicubic downsampling ~\cite{zhang2017learning} from HR images. OSAG number is set to 5, and channel number of the whole network is set to 64. The attention head number and window size are set to 4 and 8 for both Meso-OSAB and Global-OSAB. 
We use AdamW~\cite{adamw} optimizer to train the model with a batch size of 64 for 800K iterations. The initial learning rate is set to $5 \times 10^{-4}$ and halved for every 200k iterations.
In each training batch, we randomly crop LR patches of size $64\times64$ as input. Our method is implemented with PyTorch~\cite{pytorch}, and all experiments are conducted on one NVIDIA V100 GPU.
Note that no other data augmentation (e.g., Mixup~\cite{mixup}, RGB channel shuffle) or training skills (e.g., pre-training~\cite{edt}, cosine learning schedule~\cite{cosine}) are employed.
It should be pointed out that we maintain the consistency of model parameters in the ablation study by adjusting the channels of $1 \times 1$ convolution.


\subsection{Comparison with the SOTA SR methods}
To evaluate the effectiveness of Omni-SR, we compare our model with several advanced lightweight SR methods under a scale factor of 2/3/4. 
In particular, former works, VDSR~\cite{vdsr}, CARN~\cite{CARN}, IMDN
~\cite{imdn}, EDSR~\cite{edsr},  RFDN~\cite{liu2020residual}, MemNet~\cite{memnet}, MAFFSRN~\cite{muqeet2020multi}, LatticeNet~\cite{latticenet}, RLFN~\cite{RLFN}, ESRT~\cite{ESRT} and SwinIR~\cite{swinir} are introduced for comparison. 

\begin{figure}[t]	
    \centering	
	\includegraphics[width=1.0\linewidth]{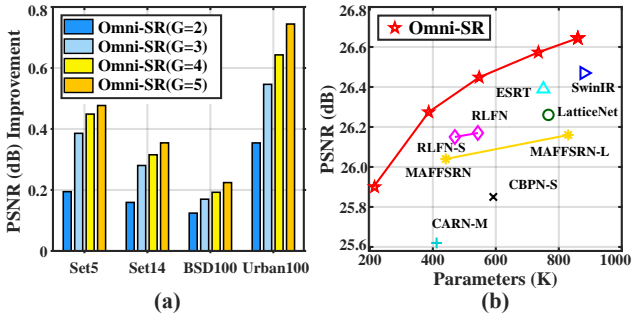}
	\vspace{-6mm}
	\caption{
    (a) PSNR improvement of Omni-SR variants with different OSAG number (K) over the smallest Omni-SR model (K=1) for 4 $\times$ SR. (b) The number of model parameters
    \emph{vs.} PSNR of different lightweight methods on Urban100 dataset for 4 $\times$ SR.}
    \label{fig:params}
    \vspace{-6mm}
\end{figure}

\begin{figure*}[!t]
\begin{center}
\scalebox{1}{
\renewcommand{\arraystretch}{0.8}
\begin{tabular}[b]{c@{ } c@{ }  c@{ } c@{ } c@{ } c@{ }	}
\hspace{-2mm}
    \multirow{4}{*}{\includegraphics[height=29.8mm,width=44.5mm,valign=t]{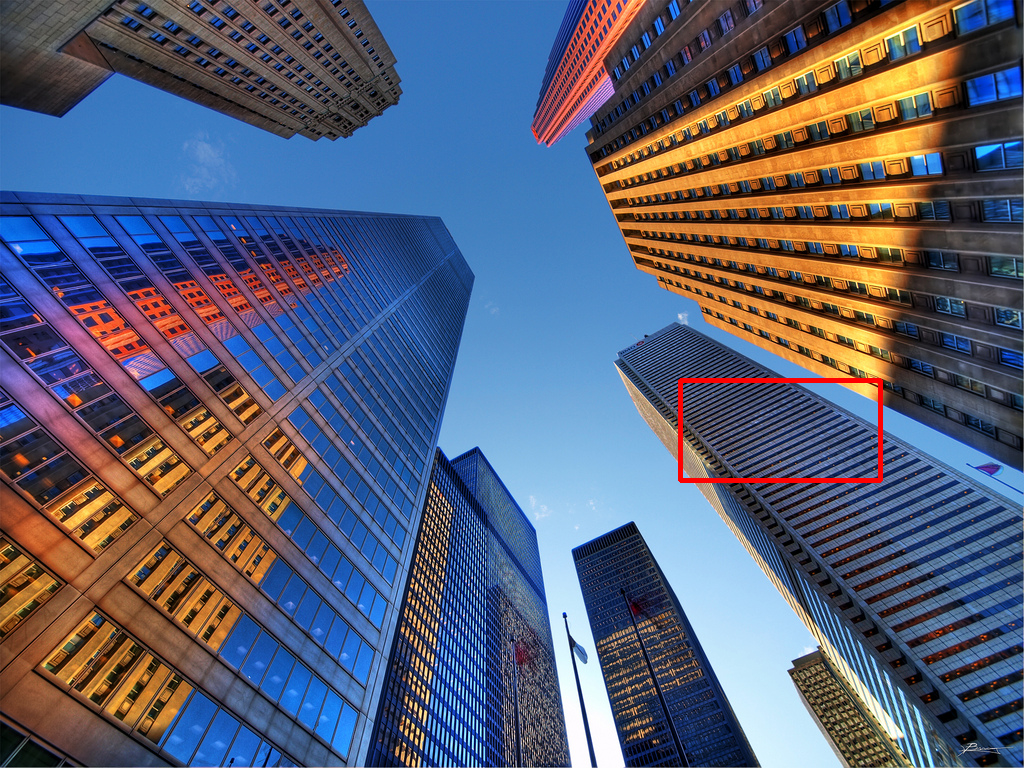}} &   
    \includegraphics[width=0.135\textwidth,valign=t]{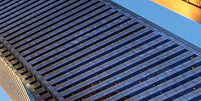}&
  	\includegraphics[width=0.135\textwidth,valign=t]{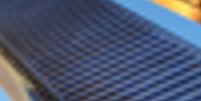}&   
      \includegraphics[width=0.135\textwidth,valign=t]{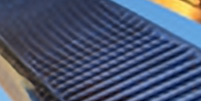}&
    \includegraphics[width=0.135\textwidth,valign=t]{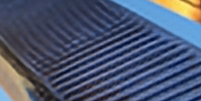}&
        \includegraphics[width=0.135\textwidth,valign=t]{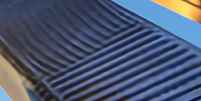}
\\
    &  \small~PSNR &\small~20.65 dB  & \small~21.15 dB & \small~21.22 dB & \small~21.99 dB   \\
    
    & \small~Reference & \small~Bicubic    & \small~SRCNN ~\cite{SRCNN}& \small~FSRCNN~\cite{lai2018fast}& \small~CARN~\cite{CARN} \\

    &

    \includegraphics[width=0.135\textwidth,valign=t,valign=t]{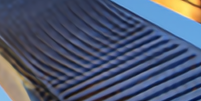}&  
     \includegraphics[width=0.135\textwidth,valign=t]{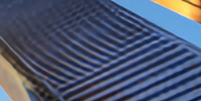}&
     \includegraphics[width=0.135\textwidth,valign=t]{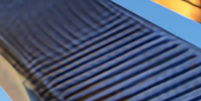}&
    \includegraphics[width=0.135\textwidth,valign=t]{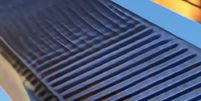}&
     \includegraphics[width=0.135\textwidth,valign=t]{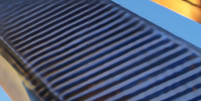}\\

     \multirow{2}{*}{Urban100: \texttt{img\_012.png}} & \small~21.95 dB & \small~21.95 dB & \small 22.00 dB  & \small 22.34 dB & \small~\textbf{24.16 dB}\\
     & \small~IMDN~\cite{imdn} & \small~RLFN ~\cite{RLFN}   & \small LatticeNet~\cite{latticenet}& \small~SwinIR~\cite{swinir} & \small~\textbf{Omni-SR (Ours)}
\\
\hspace{-2mm}
    \multirow{4}{*}{\includegraphics[height=29.8mm,width=44.5mm,valign=t]{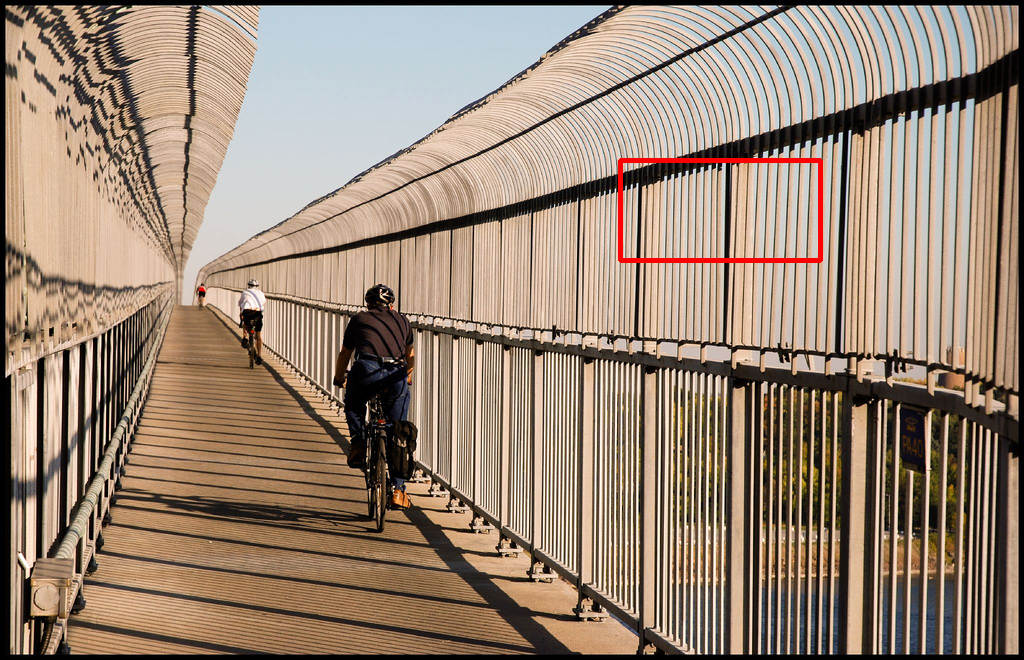}} &   
    \includegraphics[width=0.135\textwidth,valign=t]{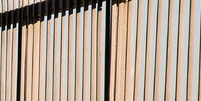}&
  	\includegraphics[width=0.135\textwidth,valign=t]{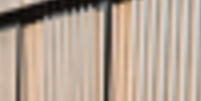}&   
      \includegraphics[width=0.135\textwidth,valign=t]{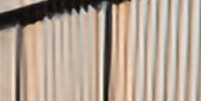}&
    \includegraphics[width=0.135\textwidth,valign=t]{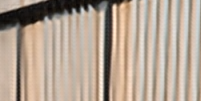}&
        \includegraphics[width=0.135\textwidth,valign=t]{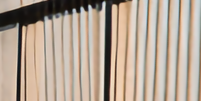}
\\
    &  \small~PSNR &\small~16.90 dB  & \small~17.54 dB & \small~17.71 dB & \small~18.78 dB   \\
    
    & \small~Reference & \small~Bicubic    & \small~SRCNN ~\cite{SRCNN}& \small~FSRCNN~\cite{lai2018fast}& \small~CARN~\cite{CARN} \\

    &

    \includegraphics[width=0.135\textwidth,valign=t]{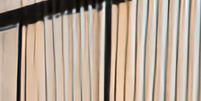}&  
     \includegraphics[width=0.135\textwidth,valign=t]{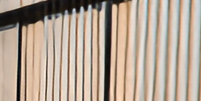}&
     \includegraphics[width=0.135\textwidth,valign=t]{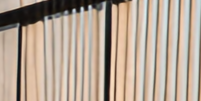}&
    \includegraphics[width=0.135\textwidth,valign=t]{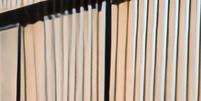}&
     \includegraphics[width=0.135\textwidth,valign=t]{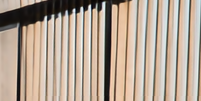}\\

    \multirow{2}{*}{Urban100: \texttt{img\_024.png}} & \small~18.77 dB & \small~19.23 dB & \small 18.91 dB  & \small 19.07 dB & \small~\textbf{21.18 dB}\\
      & \small~IMDN~\cite{imdn} & \small~RLFN ~\cite{RLFN}   & \small LatticeNet~\cite{latticenet}& \small~SwinIR~\cite{swinir} & \small~\textbf{Omni-SR (Ours)}
\\
\hspace{-2mm}
    \multirow{4}{*}{\includegraphics[height=29.8mm,width=44.5mm,valign=t]{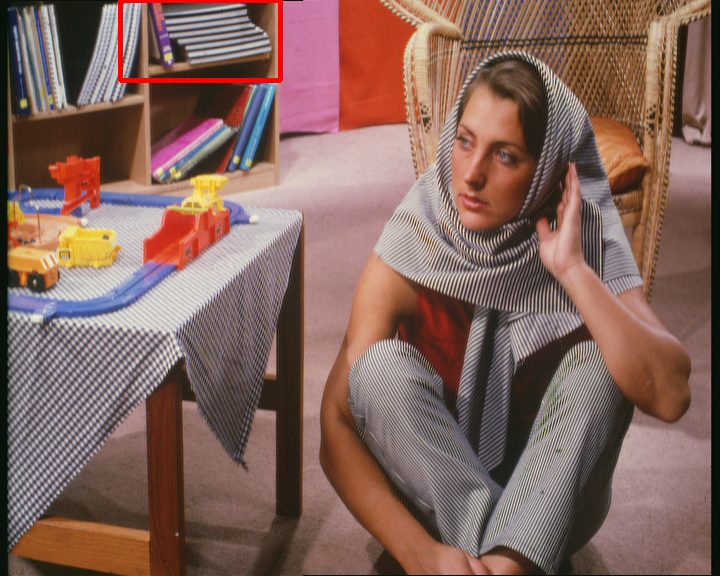}} &   
    \includegraphics[width=0.135\textwidth,valign=t]{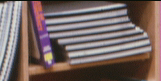}&
  	\includegraphics[width=0.135\textwidth,valign=t]{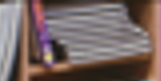}&   
      \includegraphics[width=0.135\textwidth,valign=t]{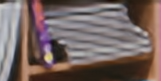}&
    \includegraphics[width=0.135\textwidth,valign=t]{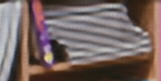}&
        \includegraphics[width=0.135\textwidth,valign=t]{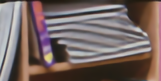}
\\
    &  \small~PSNR &\small~23.58 dB  & \small~24.04 dB & \small~24.11 dB & \small~24.48 dB   \\
    
    & \small~Reference & \small~Bicubic    & \small~SRCNN ~\cite{SRCNN}& \small~FSRCNN~\cite{lai2018fast}& \small~CARN~\cite{CARN} \\
    &

    \includegraphics[width=0.135\textwidth,valign=t]{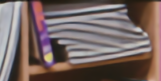}&  
     \includegraphics[width=0.135\textwidth,valign=t]{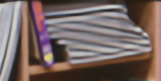}&
     \includegraphics[width=0.135\textwidth,valign=t]{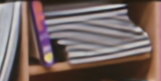}&
    \includegraphics[width=0.135\textwidth,valign=t]{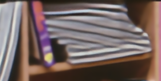}&
     \includegraphics[width=0.135\textwidth,valign=t]{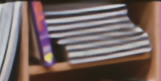}\\

     \multirow{2}{*}{Set14: \texttt{img\_barbara.png}} & \small~24.62 dB & \small~24.43 dB & \small 24.64 dB  & \small 24.45 dB & \small~\textbf{26.19 dB}\\
     & \small~IMDN~\cite{imdn} & \small~RLFN ~\cite{RLFN}   & \small LatticeNet~\cite{latticenet}& \small~SwinIR~\cite{swinir} & \small~\textbf{Omni-SR (Ours)}
\\
\end{tabular}
}
\end{center}
\vspace{-6mm}
\caption{Visual comparison for $\times4$ SR methods. The patches for comparison are marked with \red{red} boxes.
(Best viewed by zooming.)
}
\label{fig:vis}\vspace{-1em}
\end{figure*}


\textbf{Quantitative results.} In Table~\ref{tab:psnr-ssim}, the quantitative comparisons of different lightweight methods are presented on five benchmark datasets.
With a similar model size, the performance of our Omni-SR surpasses existing methods with a notable margin on all benchmarks. 
In particular, compared to other transformer architectures with comparable parameters like SwinIR~\cite{swinir} and ESRT~\cite{ESRT}, the proposed Omni-SR obtains the best performance. The results showcase that the omni-axis (i.e., spatial + channel) interaction introduced by OSA can effectively boost the model's contextual aggregation capabilities, which promises superior SR performance. Coupling with large training dataset DF2K, the performance can be further improved, especially on Urban100. We suppose that such a phenomenon can be ascribed to the images in Urban100 have many similar patches, and the long-term relationship introduced by OSAG can bring great benefits for detail restoration. More importantly, with similar parameters, our model reduces $28\%$ of computational complexity (Omni-SR: 36G FLOPs \emph{vs.} SwinIR: 50G FLOPs @1280$\times$720), showing 
its effectiveness and efficiency.




\textbf{Visual comparison.} In Figure~\ref{fig:vis}, we also provide a visual comparison of different lightweight SR methods at $\times 4$ scale. 
We can observe that the HR images constructed by Omni-SR contain more fine-grained details, while other methods generate blurred edges or artifacts in complicated areas. 
For instance, in the $1st$ row, our model is able to restore the detailed texture of the wall pleasantly, which all other methods fail to restore.
Visual results also validate the effectiveness of the proposed OSA paradigm, which can perform omni-axis pixel-wise interaction modeling, thus obtaining a more powerful reconstruction capability.

\textbf{Trade-off between Model Size and Performance.} 
In experiments, we set the number of OSAG as 5 to make the model size around 800K for a fair comparison with other methods. We also explore model performance with smaller parameter sizes by reducing OSAG number K. 
As shown in Figure~\ref{fig:params}(a), compared to the smallest model variant with K=1, increasing the number of OSAG leads to stable performance improvements.
In Figure~\ref{fig:params}(b), we present PSNR \emph{vs.} parameters of different methods. It can be found that Omni-SR achieves the best results under various settings, showing its effectiveness and scalability.

\subsection{Analysis of Omni Self-Attention}\label{sec_optimization}
\begin{figure*}[t]	
    \centering	
	\includegraphics[width=0.85\linewidth]{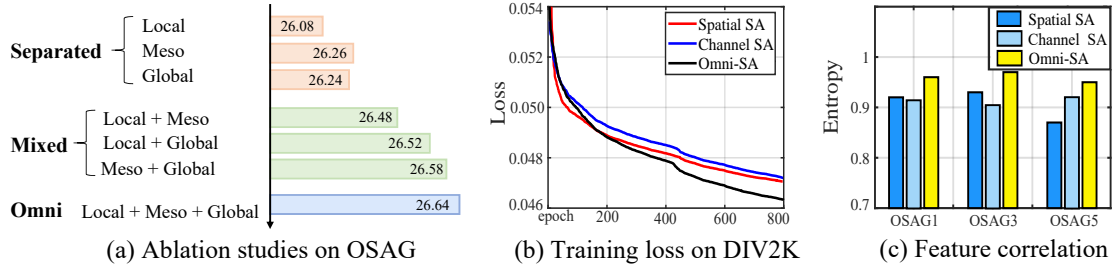}
    \vspace{-1mm}
	\caption{ (a) Ablation studies on different components in OSAG. (b) Training loss of different attention paradigms. (c) Feature correlation analysis of different attention paradigms. All the results are reported on DIV2K dataset for 4$\times$ SR. 
	}
    \label{fig:analysis}
    \vspace{-2mm}
\end{figure*}

\begin{figure}[t]
\begin{center}
\includegraphics[width=0.88\linewidth]{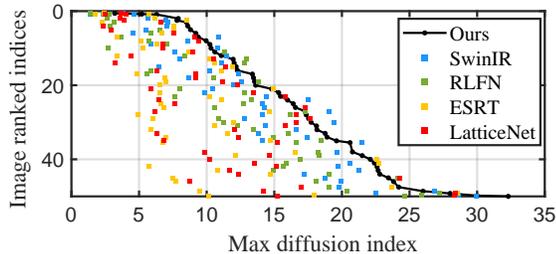}
\end{center}
\vspace{-4mm}
   \caption{The distribution of DI values for different methods on Urban100. Top-left points with show a narrow area of interest, and right-bottom points show a large area of interest.}
\label{fig:DI}
\vspace{-4mm}
\end{figure}

In this section, we illustrate the optimization features of OSA and further uncover its underlying mechanism.
Self-attention is a low-bias operator, which makes its optimization difficult and requires more training epochs.
For this, we introduce the additional channel-wise interaction to alleviate it.
In Figure~\ref{fig:analysis}(a), 
we show the loss curves of different self-attention paradigms on the DIV2K training set, including spatial self-attention, channel self-attention, and the proposed omni self-attention.
We can see that our OSA presents a obviously superior convergence speed.
More importantly, the performance at the final epoch is also significantly ahead of them.
The above phenomenon clearly shows that our OSA has superior good optimization characteristics.
Further, we delve into why channel-wise interactions lead to these improvements.
We calculate the normalized entropies~\cite{DBLP:conf/cvpr/WangLV21} of the hidden layer features of the network composed of the above three computational primitives.
We illustrate the entropy results in Figure~\ref{fig:analysis}(c).
As shown in the figure, in all outgoing layers, our OSA-encoded features show higher entropy, indicating that our OSA encodes richer information.
More information may come from various scales, and this information can help the operators to reconstruct the exact details faster.
We speculate that this is the potential reason why our OSA shows better optimization properties. In addition, following previous works~\cite{DBLP:conf/cvpr/GuD21,HAT}, we also resort to LAM analysis. DI~\cite{DBLP:conf/cvpr/GuD21} metric can measure the furthest interaction distance of the model. From Figure~\ref{fig:DI} we can observe that Omni-SR generally has the highest max diffusion index than other methods, showing that our OSA paradigm can effectively capture long-range interactions.

\subsection{Ablation Study}
\label{sec_ablation}
\textbf{Effect of Omni Self-Attention.}
The core idea of our framework is to extend the vanilla self-attention with a channel-wise relationship to build omni-axis pixel-wise interaction. Based on Omni-SR framework, we design several variant models, and their SR results are reported in Table~\ref{tab:ablation}. 
We first simply remove the channel-wise component to form a spatial-only variant (Omni-SR$_\textrm{sp}$), and its performance is degraded by $0.13$dB compared to the full model. Such a significant degradation justifies the importance of channel interactions.
Note that Omni-SR$_\textrm{sp}$ still outperforms SwinIR by 0.04dB@Urban100 $\times 4$, which benefits from the global interaction introduced by grid window partition.
Similarly, we remove the spatial self-attention component to derive a channel self-attention variant, Omni-SR$_\textrm{ca}$, and such a modification also leads to undesirable performance degradation.
Besides, we use the most widely adopted channel and spatial attention configurations (i.e., SE~\cite{hu2018squeeze} and CBAM~\cite{woo2018cbam}) to act as alternative operators for channel and spatial aggregation.
Both substitutions (Omni-SR$_{\mathrm{SE}}$, Omni-SR$_{\mathrm{{\small CBAM}}}$) hurt PNSR performance compared to the full model.
The above results show that the specific interaction paradigm (e.g., scalar-based, covariance-based) is equally important, and our channel interaction based on the covariance matrix shows great advantages.

\textbf{Effect of Omni-Scale Aggregation Group.}
In Omni-SR, we propose a local-meso-global interaction scheme (i.e., OSAG) to pursue progressive feature aggregation.
To investigate its effectiveness, we design three different kinds of interaction schemes based on Omni-SR framework: \textbf{Separated} scheme, \textbf{Mixed} scheme, and our fully designed \textbf{Omni} scheme (i.e., our proposed OSAG), and the ablation study results are shown in Figure~\ref{fig:analysis}(a).
In the figure, we employ different words (e.g., ``Local", ``Meso+Global") to represent specific schemes, e.g., ``Local" denotes using Local-Conv block to replace Meso-OSA and Global-OSA; ``Local+Global" represents replace original cascaded Meso-OSA and Global-OSA with cascaded Local-Conv and Global-OSA.
We can observe that single interaction schemes (e.g., ``Local") perform the worst.
Interestingly, the ``Global" scheme is inferior to the ``Meso" one due to its poor optimization properties of global self-attention~\cite{halo,aaconv,swin}.
Once two interaction operators are combined, the performance improves steadily.
Among them, ``Meso+ Global" setting achieves the second-best performance.
Furthermore, combining all three interaction schemes together, we obtain the best performing scheme, i.e., ``Omni".
From the above experiments, we can infer that obvious performance gains can be obtained by introducing various-scale interactions, which also illustrates the feasibility and effectiveness of our proposed OSAG.

\begin{table}[t]
\centering
\caption{Ablation studies of omni self-attention on Urban100. Omni-SR$_{(*)}$ denotes different modifications. `SA' and `S-SA' denote spatial gate and spatial self-attention. `CA' and `C-SA' denote channel gate and channel self-attention. We maintain the consistency of model parameters by adjusting the channels of $1 \times 1 Conv$.}
\label{tab:ablation}
\vspace{-2mm}
\small
\begin{spacing}{0.95}
\setlength{\tabcolsep}{0.5mm}
\begin{tabular}{l|cccc|c|ccc}
\bottomrule[0.8pt]
 \multicolumn{1}{l|}{Model}& 
  \multicolumn{1}{c}{SA}&
 \multicolumn{1}{c}{S-SA}&
  \multicolumn{1}{c}{CA}&
  \multicolumn{1}{c|}{C-SA}&
  \multicolumn{1}{l|}{FLOPs}&
\multicolumn{1}{c}{$\times$2}&
\multicolumn{1}{c}{$\times$3}&
\multicolumn{1}{c}{$\times$4}\\\hline 
Omni-SR$_{\mathrm{sp}}$ &\xmarkg&\checkmark &\xmarkg&\xmarkg& 33G&  32.88 & 28.72 & 26.51 \\
Omni-SR$_{\mathrm{SE}}$ &\xmarkg&\checkmark &\checkmark & \xmarkg& 34G& 32.83 & 28.71 & 26.50 \\
Omni-SR$_{\mathrm{{\small CBAM}}}$ &\checkmark&\xmarkg &\checkmark & \checkmark& 34G& 32.92 & 28.76 & 26.53 \\
Omni-SR$_{\mathrm{{\small ca}}}$ &\xmarkg&\xmarkg &\xmarkg&\checkmark& 33G& 32.65 & 28.60 & 26.45 \\
\rowcolor{gray!25}Omni-SR$_{\textrm{full}}$ &\xmarkg&\checkmark&\xmarkg&\checkmark&36G& 33.05 & 28.84 & 26.64 
\\\bottomrule[0.8pt]
\end{tabular}
\end{spacing}
\vspace{-2mm}
\end{table}

\section{Conclusion}
\label{sec:conclusion}
In this work, we propose Omni-SR, a lightweight framework for image SR. We propose the Omni Self-attention paradigm for simultaneous spatial and channel interactions, mining all the potential correlations across omni-axis. Furthermore, we propose an omni-scale aggregation scheme to effectively enlarge the receptive fields with low computational complexity, 
which encodes contextual relations in a progressively hierarchical manner.
Extensive experiments on public benchmark datasets and comprehensive analytical studies validate its prominent SR performance. 

\vspace{-0.5mm}
\section{Acknowledgement} \label{sec7}
This work was supported by National Science Foundation of China (U20B2072, 61976137). This work was also partly supported by SJTU Medical Engineering Cross Research Grant YG2021ZD18.

\clearpage
{\small
\bibliographystyle{ieee_fullname}
\bibliography{egbib}
}

\end{document}